\def\year{2021}\relax
\documentclass[letterpaper]{article} 
\usepackage{aaai21}  
\usepackage{times}  
\usepackage{helvet} 
\usepackage{courier}  
\usepackage[hyphens]{url}  
\usepackage{graphicx} 
\usepackage{natbib}  
\usepackage{caption} 
\usepackage{graphicx}
\nocopyright
\usepackage{color}
\usepackage{amsmath}
\usepackage{booktabs}

\usepackage{listings}

\title{GPT3-to-plan: Extracting plans from text using GPT-3}
\author{
    Alberto Olmo,
    Sarath Sreedharan,
    Subbarao Kambhampati
    \\
}
\affiliations{
    Arizona State University\\
    \{aolmoher, ssreedh3, rao\}@asu.edu
}

\begin{document}

\maketitle

\begin{abstract}
Operations in many essential industries including finance and banking are often characterized by the need to perform repetitive sequential tasks. Despite their criticality to the business, workflows are rarely fully automated or even formally specified, though there may exist a number of natural language documents describing these procedures for the employees of the company. Plan extraction methods provide us with the possibility of extracting structure plans from such natural language descriptions of the plans/workflows, which could then be leveraged by an automated system. In this paper, we investigate the utility of generalized language models in performing such extractions directly from such texts. Such models have already been shown to be quite effective in multiple translation tasks, and our initial results seem to point to their effectiveness also in the context of plan extractions. Particularly, we show that GPT-3 is able to generate plan extraction results that are comparable to many of the current state of the art plan extraction methods.
\end{abstract}

\section{Introduction}
Following sequential procedures and plans undergird many aspects of our everyday lives. As we look at many vital and consequential industries, including finance and banking, the ability to identify the correct procedures and adhere to them perfectly, becomes essential. So it is of no surprise that many enterprises invest heavily in accurately documenting these workflows in forms that are easy for their employees to follow. As we start automating many of these day-to-day activities, it becomes important that our automated systems are also able to pick up and execute them. Unfortunately, having these procedures documented is not the same as them being easy and readily available for an AI system to use. Additionally, in many of these high-risk domains, the agent cannot just try to figure out these procedures on their own through trial and error. Instead, we would want to develop ways wherein we can convert these procedures designed for human consumption to easier forms for agents to use. Within the planning community, there has been a lot of recent interest in developing {\em plan extraction} methods that are able to take natural language text describing a sequential plan. Some of the more recent works in this direction include, works like \citet{easdrl, daniele2017navigational}, which have proposed specialized frameworks for performing sequence-to-sequence translation that maps natural language sentences into structured plans.

On the other hand, the mainstream Natural Language Processing (NLP) has started shifting its focus from more specialized translation methodologies to developing general purpose models such as transformer networks \cite{radfordGPT2, brownGPT3}. These networks have already shown very encouraging results in many tasks and proven their ability to generalize to unseen ones. These are task-agnostic language models trained on large general web corpora and have shown to be comparable (and in some cases better than) their state-of-art task-specific counterparts. Examples of some tasks these models have been tested on includes, question-answering, translation, on-the-fly reasoning and even generation of news articles that are arguably indistinguishable from human-written ones. In light of these advancements, we try to answer the following question: \textit{to what extent can the current state-of-art in general natural language models compete against task-specific action sequences extractors?} 
These papers have generally looked at employing learning based methods that expect access to large amounts of pre-processed/task-specific data, including annotations that allow mapping of text to the required structured output. 
These characteristics make the methods fragile to changes in input and output format. Combining this with the need for extensive training data, we expect these systems 
to require heavy time and resource investment and expert oversight to set up.

In this paper, we want to investigate how GPT-3 \cite{brownGPT3}, one of the most recent transformer-based language models, can be used to extract structured actions from natural language texts. We find that these models achieve comparable, and in some cases better scores than previous state-of-the-art task specific methods. 
We make use of natural language text from three domains and measure the performance of the model in terms of its $F_1$ score, a commonly used quantitative measure for the task. We then compare it to previously published results for task-specific action extractors which use a varied range of solutions, including, reinforcement learning, \cite{easdrl}, sequence-to-sequence models \cite{daniele2017navigational}, Bi-directional LSTMs \cite{ma2016blcc} or clustering of action templates \cite{lindsay2017StanfordCore}.
 
The proliferation and effectiveness of such general language models even in specific tasks, open up new opportunities for planning researchers and practitioners. In particular, it empowers us to deploy planning techniques in real-world applications without worrying about the natural-language interaction aspects of the problem. Also, note that all results reported here are directly calculated from the best GPT-3 raw predictions, with no additional filtering or reasoning employed atop of it. We expect most of the results reported here to improve should we additionally exploit domain-level or task-level insights to filter the results from these models. 

\section{Background and Related Works}
The Generative Pre-trained Transformer 3 (GPT-3) \cite{brownGPT3} is the latest version of the GPT models developed by OpenAI\footnote{https://openai.com/}. A 175 billion parameter autoregressive language model with 96 layers trained on a 560GB+ web corpora (Common Crawl\footnote{https://commoncrawl.org/} and WebText2 \cite{OpenWebText}), internet-based book corpora and Wikipedia datasets each with different weightings in the training mix and billions of tokens or words. Tested on several unrelated natural language tasks, GPT-3 has proven successful in generalizing to them with just a few examples (zero in some cases). GPT-3 comes in 4 versions, Davinci, Curie, Babbage and Ada which differ in the amount of trainable parameters -- 175, 13, 6.7 and 2.7 billion respectively \cite{brownGPT3}.
Previous work on action sequence extraction from descriptions has revolved around specific models for action extraction, some of them trained on largely task-specific preprocessed data. \cite{mei2016listen, daniele2017navigational} use sequence-to-sequence models and inverse reinforcement learning to generate instructions from natural language corpora. Similarly, \citet{easdrl} uses a reinforcement learning model to extract word actions directly from free text (i.e. the set of possible actions is not provided in advance) where, within the RL framework, actions select or eliminate words in the text and states represent the text associated with them. This allows them to learn the policy of extracting actions and plans from labeled text. In a same fashion, \citet{rlBranavan} also use Reinforcement Learning, a policy gradient algorithm and a log-linear model to predict, construct and ultimately learn the sequence of actions from text. Other works like \citet{unstructuredBorrajo} define a system of tools through which they crawl, extract and denoise data from plan-rich websites and parse their actions and respective arguments with statistical correlation tools to acquire domain knowledge.

However, to the best of our knowledge this paper is the first work to assess the performance of a general purpose NLP language model on action sequence extraction tasks compared to its current state-of-art task-specific counterpart. 


\section{Experiments}

\begin{table}[]
\begin{center}
\begin{tabular}{lccc}
\toprule
 & WHS & CT & WHG \\ \midrule
Labeled texts & 154 & 116 & 150 \\
Input-output pairs & 1.5K & 134K & 34M \\
Action name rate (\%) & 19.47 & 10.37 & 7.61 \\
Action argument rate (\%) & 15.45 & 7.44 & 6.30 \\
Unlabeled texts & 0 & 0 & 80 \\ \bottomrule
\end{tabular}   
\end{center}
\caption{Characteristics of the datasets used.}
\label{tab:datasets}
\vspace{-5pt}
\end{table}

\paragraph{Datasets and GPT-3 API} We use the three most common datasets for action sequence extraction tasks used in evaluating many of the previous task-specific approaches, including \citet{easdrl} or \citet{miglani2020nltopddl}. Namely, the "Microsoft Windows Help and Support" (WHS), the "WikiHow Home and Garden" (WHG) and the "CookingTutorial" (CT) datasets. The characteristics of these datasets are provided in Table \ref{tab:datasets}. 

The GPT-3 model is currently hosted online\footnote{More information at https://beta.openai.com/} and can be accessed via paid user queries with either their API or website in real time. Some example use cases of their service include keyword extraction from natural text, mood extraction from reviews, open-ended chat conversations and even text to SQL and JavaScript to Python converters amongst many others. In general, the service takes free natural language as input and the user is expected to encode the type of interaction/output desired in the input query. The system then generates output as a completion of the provided query. The API also allows the user to further tweak the output by manipulating the following parameters: \texttt{Max Tokens} sets the maximum number of words that the model will generate as a response, \texttt{Temperature}  (between 0 and 1) allows the user to control the randomness (with 0 forcing the system to generate output with the highest probability consistently and rendering it effectively deterministic for a given input). \texttt{Top P} also controls diversity; closer to 1 ensures more determinism, \texttt{Frequency Penalty} and \texttt{Presence Penalty} penalize newly generated words based on their existing frequency so far, and \texttt{Best of} is the number of multiple completions to compute in parallel. It outputs only the best according to the model. In Table \ref{tab:gpt3-params} we show the values that we used for all our experiments to ensure the most consistency in the model's responses.

\begin{table}[]
\begin{center}
\begin{tabular}{llllll}
\toprule
Length & Temp. & Top P & Freq. & Pres. & Best of \\
\midrule
\multicolumn{1}{c}{100} & \multicolumn{1}{c}{0.0} & \multicolumn{1}{c}{1} & \multicolumn{1}{c}{0.0} & \multicolumn{1}{c}{0.0} & \multicolumn{1}{c}{1}
\end{tabular} 
\end{center}
\caption{GPT-3 parameters used for all our experiments. }
\label{tab:gpt3-params}
\vspace{-10pt}
\end{table}

\vspace{-10pt}
\paragraph{Query generation}
Each query consists of a few shot training in natural language text and the corresponding structured representation of the plan. For each example, we annotate the beginning of the natural language text portion with the tag \texttt{TEXT} followed by the plan (annotated with the tag \texttt{ACTIONS}). In the structure representation, each action is represented in a functional notation of the form $a_0^j(arg_0^0, arg_1^0 \dots arg_k^0) \dots \ a_n^j(arg_0^n, arg_1^n \dots arg_k^n)$ where $a_i^j$ represents action \textit{i} in sentence $j$ and $arg_k^n$ is the \textit{kth} argument from action $a_n$ in the text. 
After the training pairs, we include the test sample in natural language text after another tag \texttt{TEXT} and then we add a final tag \texttt{ACTIONS}, with the expectation that GPT3 will generate the corresponding plan representation after that. 

\vspace{-10pt}
\paragraph{Evaluation and Metrics} In order to directly compare the performance of GPT-3 to \citet{miglani2020nltopddl}, the current state-of-art, we followed a translation scheme with three types of actions, namely, \textit{essential} (essential action and its corresponding arguments should be included in the plan) \textit{exclusive} (the plan must only contain one of the exclusive actions) and \textit{optional} actions (the action may or may not be part of the plan). We use this scheme to generate both the example data points provided to the system and to calculate the final metrics. 

In particular, we will use \textit{precision}, \textit{recall} and \textit{F1}, similar to \citet{easdrl, miglani2020nltopddl} to measure the effectiveness of the method.

\begin{equation}
\begin{split}
    Precision &= \frac{\#TotalRight}{\#TotalTagged},\ Recall=\frac{\#TotalRight}{\#TotalTruth}\\
    \vspace{2em}
    F_1&=\frac{2\times precision \times recall}{precision + recall}
\end{split}
\label{prec_recall_f1}
\end{equation}

Note that the ground truth number and the number of true extracted actions depend on the type that each action in the text corresponds to. For example, a set of exclusive actions only contribute one action to \textit{\#TotalTruth} and we only count an extracted exclusive action in \textit{\#TotalRight}, if and only if, one of the exclusive actions is extracted. Both \textit{essential} and \textit{optional} actions only contribute once to \textit{\#TotalTruth} and \textit{\#TotalRight}. 

\vspace{-10pt}
\paragraph{Baselines} In Table \ref{tab:results} we compare GPT-3 to several action sequence extractor models:
\begin{itemize}
    \item EAD: \citet{mei2016listen} design an Encoder-Aligner-Decoder method that uses a neural sequence-to-sequence model to translate natural language instructions into action sequences.
    \item BLCC: The Bi-directional LSTM-CNN-CRF model from \citet{ma2016blcc} benefits from both word and character-level semantics and implement an end-to-end system that can be applied to action sequence extraction tasks with pre-trained word embeddings.  
    \item Stanford CoreNLP: in \citet{lindsay2017StanfordCore} they reduce Natural Language texts to action templates and based on their functional similarity, cluster them and induce their PDDL domain using a model acquisition tool.
    \item EASDRL and cEASDRL: \citet{easdrl} and \citet{miglani2020nltopddl} use similar reinforcement learning approaches; they define two Deep Q-Networks which perform the actions of \textit{selecting} or \textit{rejecting} a word. The first DQN handles the extraction of Essential, Exclusive and Optional actions while the second uses them to select and extract relevant arguments.
\end{itemize}
The corresponding \textit{precision}, \textit{recall} and \textit{F1} scores for each method were picked directly from their respective papers.
\begin{table}[t]
\centering
\large
\tabcolsep=0.2cm
\scalebox{0.85}{
\begin{tabular}{lcccccc}
\toprule
 & \multicolumn{3}{c}{\textit{Action names}} & \multicolumn{3}{c}{\textit{Action arguments}} \\
Model & \multicolumn{1}{c}{WHS} & \multicolumn{1}{c}{CT} & \multicolumn{1}{c}{WHG} & \multicolumn{1}{c}{WHS} & \multicolumn{1}{c}{CT} & \multicolumn{1}{c}{WHG} \\ 
\midrule
EAD & 86.25 & 64.74 & 53.49 & 57.71 & 51.77 & 37.70 \\
CMLP & 83.15 & 83.00 & 67.36 & 47.29 & 34.14 & 32.54 \\
BLCC & 90.16 & 80.50 & 69.46 & 93.30 & \textbf{76.33} & 70.32 \\
STFC & 62.66 & 67.39 & 62.75 & 38.79 & 43.31 & 42.75 \\
EASDRL & 93.46 & 84.18 & 75.40 & \textbf{95.07} & 74.80 & 75.02 \\
cEASDRL & \textbf{97.32} & \textbf{89.18} & \textbf{82.59} & 92.78 & 75.81 & \textbf{76.99} \\ 
\midrule
GPT-3 &  &  &  &  &  &  \\ 
\midrule
Davinci & \textbf{86.32} & \textbf{58.14} & \textbf{43.36} & 22.90 & \textbf{29.63} & \textbf{22.25} \\
Curie & 75.80 & 35.57 & 22.41 & \textbf{31.75} & 22.16 & 13.79 \\
Babbage & 62.59 & 20.62 & 14.95 & 22.91 & 12.59 & 7.33 \\
Ada & 60.68 & 14.68 & 8.90 & 17.91 & 4.13 & 2.27 \\ 
\bottomrule
\end{tabular}
}
\caption{$F_{1}$ scores for all actions and their arguments accross the WHS, CT and WHG datasets for the state-of-art sequence extraction models and GPT-3. State-of-art task-specific model $F_1$ scores are extracted from \citet{miglani2020nltopddl, easdrl} and represent their best possible recorded performance.}
\label{tab:results}
\vspace{-10pt}
\end{table}

\vspace{-10pt}
\paragraph{Results}
Given that GPT-3 is a few-shot learner we want to know how it performs given different amounts of training samples. To measure this, we query the language model with increasing numbers of examples (with a maximum of four examples) for all domains and report their F1 scores. We stop at the four-shot mark as the total amount of \textit{tokens} or words that the request can contain is 2048. Additionally for the CookingTutorial and Wikihow Garden and Home datasets, 4-shot training examples already exceed this threshold, so we limit the length of input text to 10 sentences per training example.
Specifically, we select the training examples as 1-shot (one datapoint is selected at random from the dataset),
2-shot (the two datapoints with the largest proportion of \textit{optional} and \textit{exclusive} actions from the dataset are selected), 3-shot (the three datapoints with the largest proportion of \textit{optional}, \textit{exclusive} and \textit{essential} actions) and 4-shot (an additional random datapoint is added to 3-shot).


\begin{figure}[t]
\includegraphics[width=0.78\columnwidth]{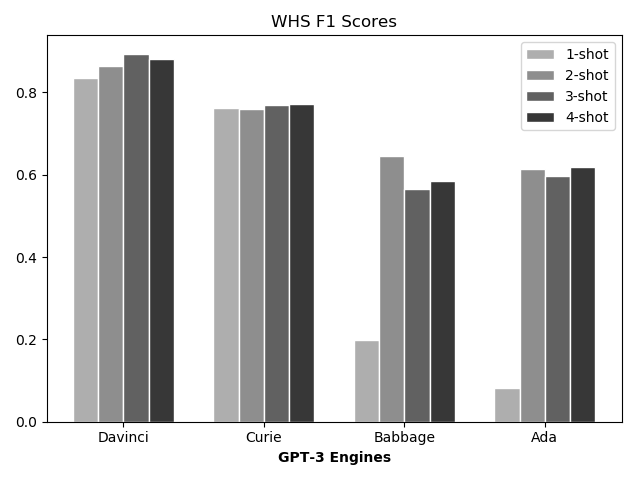}
\centering
\caption{$F_{1}$ scores of the model on the Windows Help and Support dataset for 1 to 4 few-shot training}
\label{fig:engines-few-shot}
\vspace{-5pt}
\end{figure}

In Figure \ref{fig:engines-few-shot} we show how the $F_1$ score changes given 1, 2, 3 and 4-shot training samples when tested on the whole Windows Help and Support dataset. Unsurprisingly, Davinci, the model with the most amount of trainable parameters, performs best with over $80\%\ F_1$ score for each category. Both Davinci and Curie show the tendency to perform better the more examples they are given peaking at 3 and 4-shots respectively. Similarly, Babbage and Ada show their peaks given 2 and 4 examples while underperforming at one-shot training. This is unsurprising, given the fact that these models are simplified versions of GPT-3 which have also been trained on a smaller corpus of data for higher speed. Hence, they need more than one example to grasp the task.

In table \ref{tab:results} we compare the $F_1$ scores for action name and their argument extractions as reported by previous and current state of the art task-specific action sequence extractors against all GPT-3 engines: Davinci, Curie, Babbage and Ada, ordered from most to least powerful
The scores are calculated based on \ref{prec_recall_f1} and account for \textit{essential}, \textit{exclusive} and \textit{optional} actions and their respective arguments. All GPT-3 models are trained with two-shot examples. As expected, Davinci overall performs the best compared to the rest of engines. We can see that Davinci also outperforms the EAD, CMLP and STFC task-specific models for the Windows Help and Support domain on extracting actions. Even though it underperforms on the argument extraction task compared to the state of art, it's worth nothing that still obtains better than random extraction scoring.

\begin{figure}[t]
\centering
\includegraphics[width=\columnwidth]{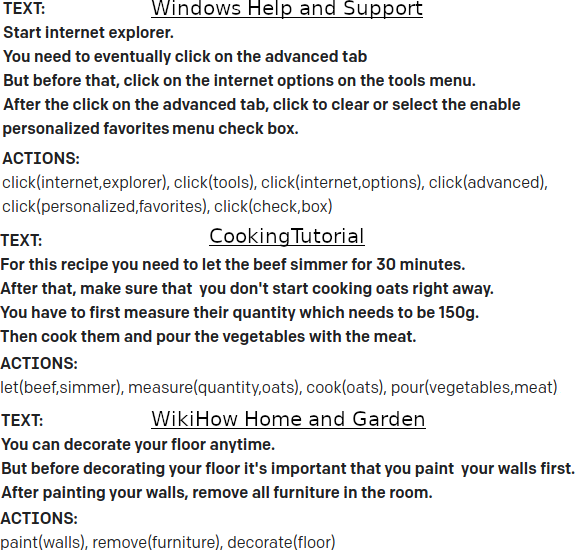}
\centering
\caption{Query examples on WHS, CT and WHG. Each query was input to Davinci along with two preceding training instances containing the largest proportion of \textit{optional} and \textit{exclusive} actions. The output is shown in regular text while the input is displayed in bold.}
\label{fig:gpt-ordering}
\vspace{-8pt}
\end{figure}

\vspace{-10pt}
\paragraph{Ordering}
We want to assess whether GPT-3 is capable of inferring plan order from text. This is a feature which is mostly missing in previous task-specific state of the art like \citet{easdrl} or \citet{miglani2020nltopddl}.
As a preliminary evaluation, we create three examples (one for each dataset, shown in Figure \ref{fig:gpt-ordering}), where order of the plans does not match how actions are listed in the text. 
In the Windows Help and Support example, we state on the second and third sentences that action \textit{click(advanced)} must be performed eventually but only after \textit{click(internet, options)}, and, even though the corresponding sentences appear in the opposite order, GPT-3 places them as expected. Similarly, in the CookingTutorial example, we state that \textit{first} we need to \textit{measure the quantity of oats} and \textit{cook them} only later and once again, it generates the actions in correct ordering. For the last example, GPT-3 shows to understand that action \textit{paint(walls)} has to be done before \textit{remove(furniture)} and, interestingly, even though \textit{decorate(floor)} is stated on the first sentence, the model seems to understand that it can be performed \textit{anytime} and places the action last. Note that these are just anecdotal evidences and we would need to perform studies over larger test sets to further evaluate GPT-3's ability to identify the ordering of plans. Our current evaluation along this dimension is limited by the lack of annotation regarding the ordering in the currently available datasets and one of our future works would be to create/identify such text-to-plan dataset with additional annotations on action ordering.

\section{Discussion and Conclusion}
In this paper we have shown that GPT-3, the state-of-art general purpose Natural Language model, can compete against task-specific approaches in the action sequence extraction domain, getting closer than ever to surpassing their performance. From the user's perspective, these transformer models pose the advantage of needing almost negligible computational resources from the user side by being readily available at just one query away and seem like a possible solution in the future to many natural language tasks should they keep up with their rate of improvement.
However, some limitations are still prevalent on GPT-3. It is still far from being accurate for the more action-diverse natural text datasets.
This becomes all the more apparent during argument extraction where, as shown, it generally fails to obtain competitive scores even on its most powerful Davinci version. Thus, this hinders the possibility of using GPT-3 directly for general extraction tasks other than the most simple. For less diverse plans, it does show competing performance and we posit that it could be used as an intermediate step in a hybrid system. 

On the other hand, GPT-3 seems to show some ability to identify the underlying sequentiality of the plan by recognizing words like  \textit{before, after, first, anytime} or \textit{eventually} and rearranging the plans accordingly. This is a capability generally missing from most state of the art plan extractors as they assume the ordering of the plan to be same as that of the sentences corresponding to each action in the text. Hence, ordering speaks for yet another potential advantage of using general models, as in they are usually not limited by specific assumptions made by system designers. Finally, note that the aforementioned strengths of the model could be further augmented should OpenAI allow for more finetuning in the future.



\section{Acknowledgements}
Dr. Kambhampati's research is supported by the J.P. Morgan Faculty Research Award, ONR grants N00014-16-1-2892, N00014-18-1-2442, N00014-18-1-2840, N00014-9-1-2119, AFOSR grant FA9550-18-1-0067 and DARPA SAIL-ON grant W911NF19-2-0006. We also want to thank OpenAI and Miles Brundage for letting us get research access to the GPT-3 API.

\bibliography{ref}

\begin{thebibliography}{11}
\providecommand{\natexlab}[1]{#1}
\providecommand{\url}[1]{\texttt{#1}}
\providecommand{\urlprefix}{URL }
\expandafter\ifx\csname urlstyle\endcsname\relax
  \providecommand{\doi}[1]{doi:\discretionary{}{}{}#1}\else
  \providecommand{\doi}{doi:\discretionary{}{}{}\begingroup
  \urlstyle{rm}\Url}\fi

\bibitem[{Addis and Borrajo(2010)}]{unstructuredBorrajo}
Addis, A.; and Borrajo, D. 2010.
\newblock From unstructured web knowledge to plan descriptions.
\newblock In \emph{Information Retrieval and Mining in Distributed
  Environments}, 41--59. Springer.

\bibitem[{Branavan et~al.(2009)Branavan, Chen, Zettlemoyer, and
  Barzilay}]{rlBranavan}
Branavan, S.; Chen, H.; Zettlemoyer, L.; and Barzilay, R. 2009.
\newblock Reinforcement Learning for Mapping Instructions to Actions.
\newblock In \emph{Proceedings of the Joint Conference of the 47th Annual
  Meeting of the {ACL} and the 4th International Joint Conference on Natural
  Language Processing of the {AFNLP}}, 82--90. Suntec, Singapore: Association
  for Computational Linguistics.
\newblock \urlprefix\url{https://www.aclweb.org/anthology/P09-1010}.

\bibitem[{Brown, Mann, and et~al.(2020)}]{brownGPT3}
Brown, T.; Mann, B.; and et~al., R. 2020.
\newblock Language Models are Few-Shot Learners.
\newblock In Larochelle, H.; Ranzato, M.; Hadsell, R.; Balcan, M.~F.; and Lin,
  H., eds., \emph{Advances in Neural Information Processing Systems},
  volume~33, 1877--1901. Curran Associates, Inc.
\newblock
  \urlprefix\url{https://proceedings.neurips.cc/paper/2020/file/1457c0d6bfcb4967418bfb8ac142f64a-Paper.pdf}.

\bibitem[{Daniele, Bansal, and Walter(2017)}]{daniele2017navigational}
Daniele, A.~F.; Bansal, M.; and Walter, M.~R. 2017.
\newblock Navigational instruction generation as inverse reinforcement learning
  with neural machine translation.
\newblock In \emph{2017 12th ACM/IEEE International Conference on Human-Robot
  Interaction (HRI}, 109--118. IEEE.

\bibitem[{Feng, Zhuo, and Kambhampati(2018)}]{easdrl}
Feng, W.; Zhuo, H.~H.; and Kambhampati, S. 2018.
\newblock Extracting action sequences from texts based on deep reinforcement
  learning.
\newblock \emph{arXiv preprint arXiv:1803.02632} .

\bibitem[{Gokaslan and Cohen(2019)}]{OpenWebText}
Gokaslan, A.; and Cohen, V. 2019.
\newblock OpenWebText Corpus.
\newblock \url{http://Skylion007.github.io/OpenWebTextCorpus}.

\bibitem[{Lindsay et~al.(2017)Lindsay, Read, Ferreira, Hayton, Porteous, and
  Gregory}]{lindsay2017StanfordCore}
Lindsay, A.; Read, J.; Ferreira, J.; Hayton, T.; Porteous, J.; and Gregory, P.
  2017.
\newblock Framer: Planning models from natural language action descriptions.
\newblock In \emph{Proceedings of the International Conference on Automated
  Planning and Scheduling}, volume~27.

\bibitem[{Ma and Hovy(2016)}]{ma2016blcc}
Ma, X.; and Hovy, E.~H. 2016.
\newblock End-to-end Sequence Labeling via Bi-directional LSTM-CNNs-CRF.
\newblock In \emph{Proceedings of the 54th Annual Meeting of the Association
  for Computational Linguistics, {ACL} 2016, August 7-12, 2016, Berlin,
  Germany, Volume 1: Long Papers}. The Association for Computer Linguistics.
\newblock \doi{10.18653/v1/p16-1101}.
\newblock \urlprefix\url{https://doi.org/10.18653/v1/p16-1101}.

\bibitem[{Mei, Bansal, and Walter(2016)}]{mei2016listen}
Mei, H.; Bansal, M.; and Walter, M. 2016.
\newblock Listen, attend, and walk: Neural mapping of navigational instructions
  to action sequences.
\newblock In \emph{Proceedings of the AAAI Conference on Artificial
  Intelligence}, volume~30.

\bibitem[{Miglani and Yorke-Smith(2020)}]{miglani2020nltopddl}
Miglani, S.; and Yorke-Smith, N. 2020.
\newblock NLtoPDDL: One-Shot Learning of PDDL Models from Natural Language
  Process Manuals.
\newblock In \emph{ICAPS'20 Workshop on Knowledge Engineering for Planning and
  Scheduling (KEPS'20)}. ICAPS.

\bibitem[{Radford et~al.(2019)Radford, Wu, Child, Luan, Amodei, and
  Sutskever}]{radfordGPT2}
Radford, A.; Wu, J.; Child, R.; Luan, D.; Amodei, D.; and Sutskever, I. 2019.
\newblock Language models are unsupervised multitask learners.
\newblock \emph{OpenAI blog} 1(8): 9.

\end{thebibliography}

\end{document}